\documentclass[runningheads]{llncs}
\usepackage[inline]{enumitem}
\usepackage{caption}
\usepackage{subcaption}
\usepackage{graphicx}
\usepackage{multirow}
\usepackage{hyperref}
\usepackage{url}
\usepackage{array}
\usepackage{tabularx}
\usepackage{ragged2e}
\usepackage{booktabs}
\usepackage{amsmath}
\usepackage{amsfonts}
\usepackage{wrapfig}
\usepackage[autolanguage]{numprint}
\usepackage{pifont}%
\usepackage{xcolor}

\newcommand{\cmark}{\ding{51}}%
\newcommand{\xmark}{\ding{55}}%

\newcommand{\codeurl}{\url{https://github.com/PaulLerner/ViQuAE}}

\newcommand{\model}{ECA}
\newcommand{\cls}{\texttt{[CLS]}}
\begin{document}
\title{Multimodal Inverse Cloze Task for Knowledge-based Visual Question Answering\thanks{This work was supported by the ANR-19-CE23-0028 MEERQAT project. This work was granted access to the HPC resources of IDRIS under the allocation 2021-AD011012846 made by GENCI.}}
\titlerunning{Multimodal Inverse Cloze Task for KVQAE}
\author{Paul Lerner\inst{1} \and
Olivier Ferret\inst{2} \and
Camille Guinaudeau\inst{1}}
\authorrunning{P. Lerner et al.}%
\institute{Université Paris-Saclay, CNRS, LISN, 91400, Orsay, France \\
\email{\{paul.lerner,camille.guinaudeau\}@lisn.upsaclay.fr}
\and
Université Paris-Saclay, CEA, List, F-91120, Palaiseau, France\\
\email{olivier.ferret@cea.fr}}

\maketitle              %

\begin{abstract}
We present a new pre-training method, Multimodal Inverse Cloze Task, for Knowledge-based Visual Question Answering about 
nam-ed
Entities (KVQAE). KVQAE is a recently introduced task that consists in answering questions about named entities grounded in a visual context using a Knowledge
Base. Therefore, the interaction between the modalities is paramount to retrieve information and must be captured with complex fusion models. As these models require a lot of training data, we design this pre-training task from existing work in textual Question Answering. It consists in considering a sentence as a pseudo-question and its context as a pseudo-relevant passage and is extended by considering images near texts in multimodal documents. Our method is applicable to different neural network architectures and leads to a 9\% relative-MRR and 15\% relative-F1 gain for retrieval and reading comprehension, respectively, over a no-pre-training baseline. %

\end{abstract}

\begin{figure}[t]
\begin{tabular}{|p{0.12\textwidth}p{0.22\textwidth}|p{0.12\textwidth}p{0.5\textwidth}|}
    \hline
    \multicolumn{2}{|c|}{\textbf{Visual Question (input)}}   &  
     \multicolumn{2}{c|}{\textbf{Relevant visual passage in the Knowledge Base}} \\
     
    \hline \raisebox{-.9\height}{\includegraphics[width=0.13\textwidth,height=0.17\textwidth]{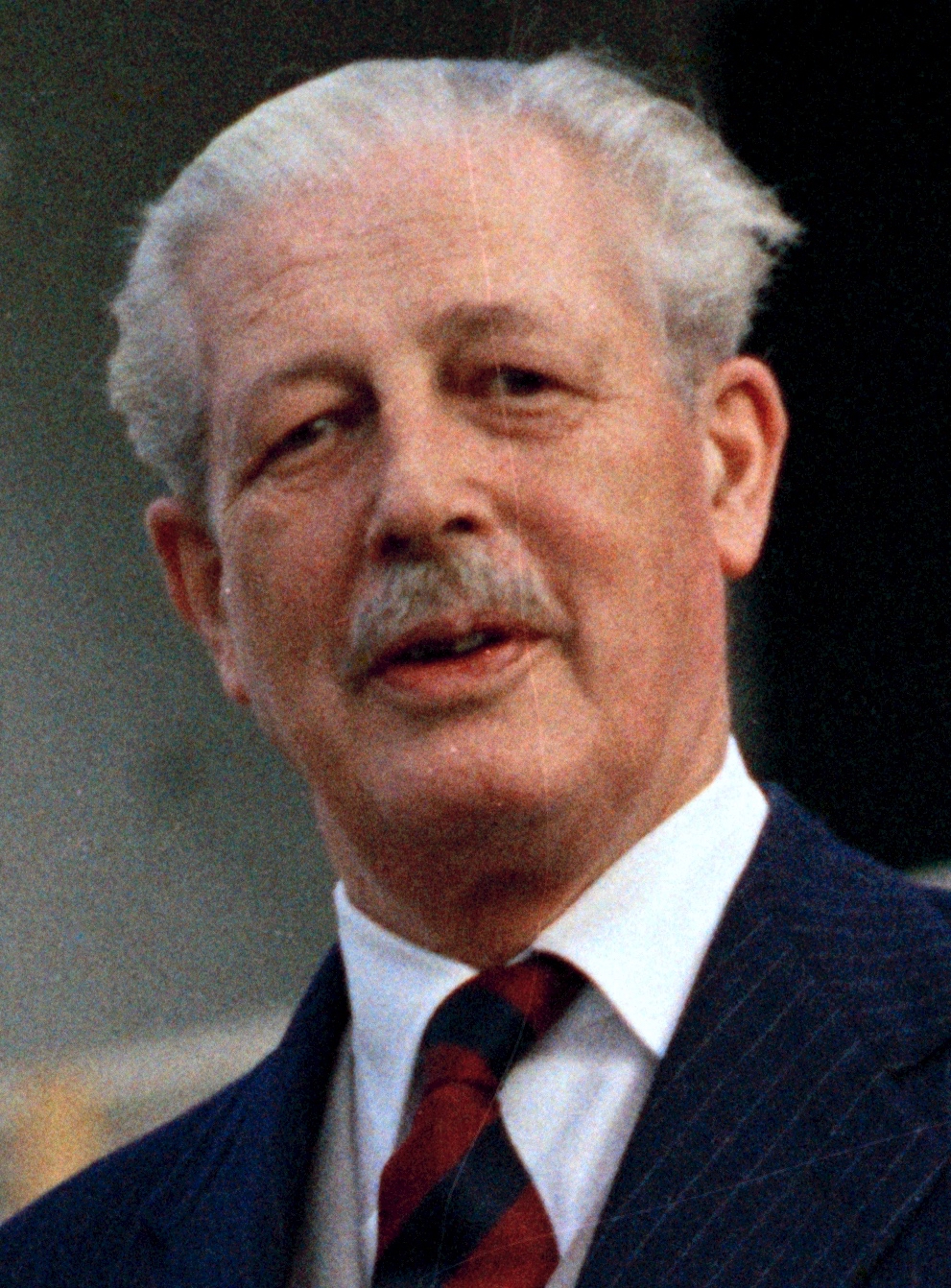}} & \scriptsize \textit{‘‘Which constituency did this man represent when he was Prime Minister?’’}  & 
    
    \raisebox{-.9\height}{\includegraphics[width=0.13\textwidth,height=0.17\textwidth]{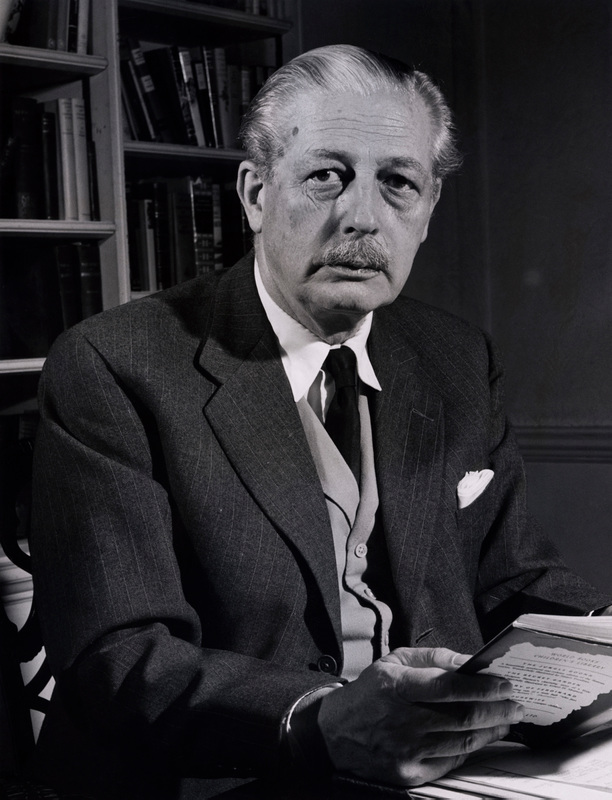}} & \scriptsize ‘‘Macmillan indeed lost Stockton in the landslide Labour victory of 1945, but returned to Parliament in the November 1945 by-election in \textbf{Bromley}.’’ \\
\hline

\centering
 \raisebox{-.9\height}{\includegraphics[width=0.13\textwidth]
 {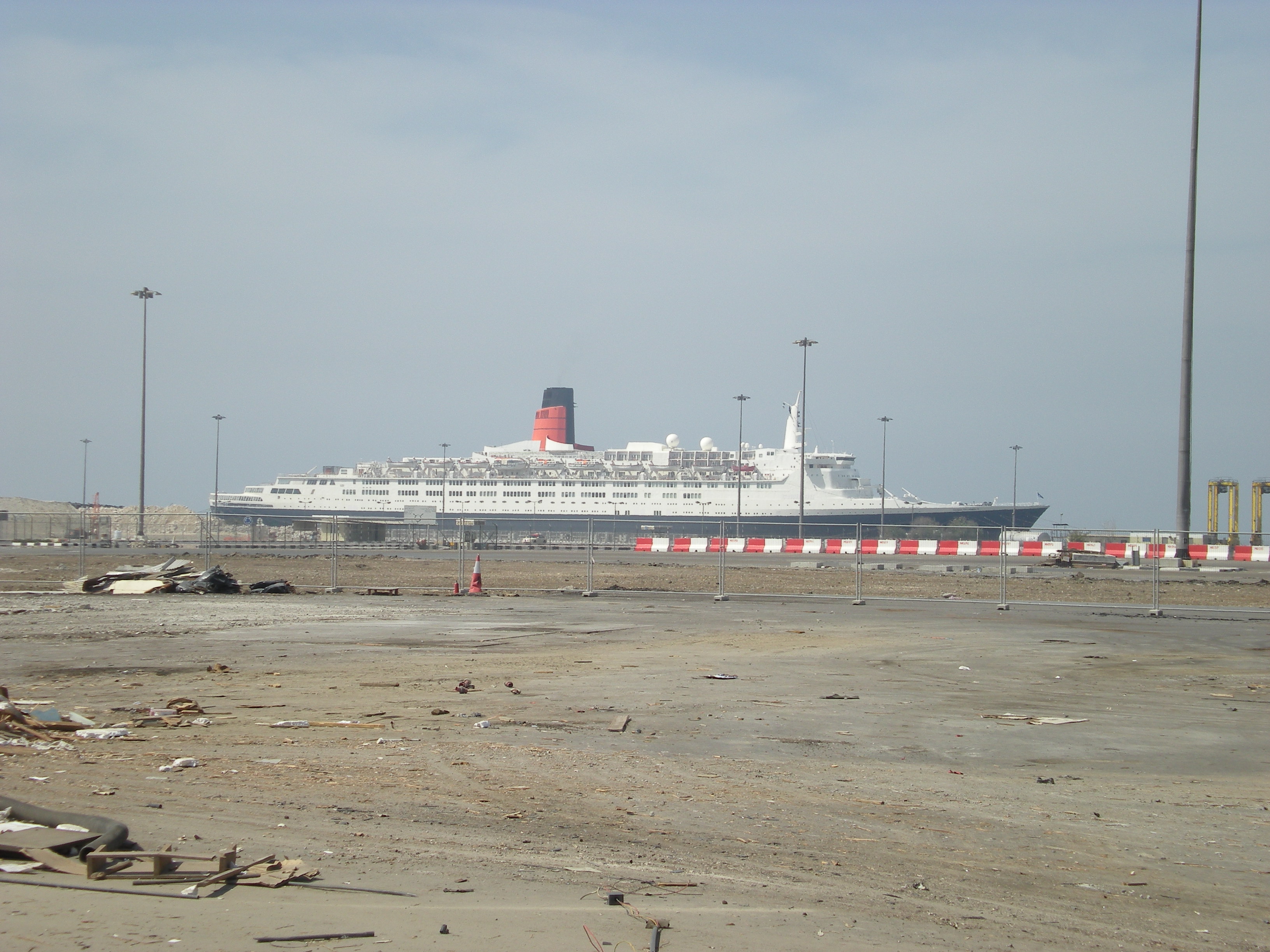}} & \scriptsize ‘‘\textit{In which year did this ocean liner make her maiden voyage?}’’ &
 \raisebox{-.9\height}{\includegraphics[width=0.13\textwidth]{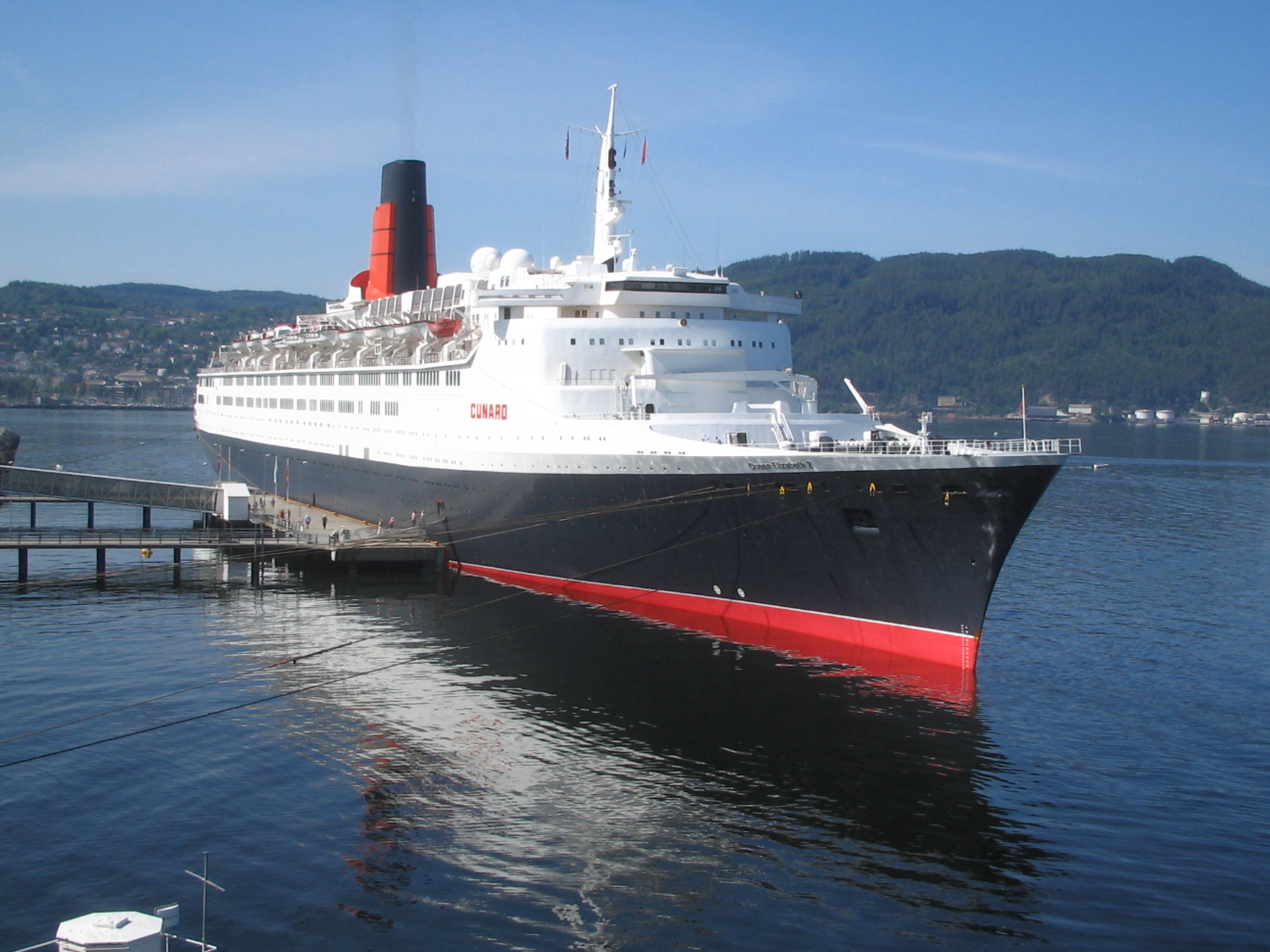}} 
 & \scriptsize ‘‘Queen Elizabeth 2, often referred to simply as QE2, is a floating hotel and retired ocean liner built for the Cunard Line which was operated by Cunard as both a transatlantic liner and a cruise ship from \textbf{1969} to 2008.’’\\

\hline
    \end{tabular}
    \caption{Example of visual questions about named entities from the ViQuAE dataset along with relevant visual passages from its Knowledge Base~\cite{sigir2022}.}
    \label{tab:kvqae_eg}
\end{figure}

\section{Introduction}\label{sec:Introduction}

Knowledge-based Visual Question Answering about named Entities (KVQAE) is a challenging task recently introduced in~\cite{shah2019kvqa}. It consists in answering questions about named entities grounded in a visual context  using a Knowledge Base (KB). Figure~\ref{tab:kvqae_eg} provides two examples of visual questions along with relevant visual passages from a KB. To address the task, one must thus \textit{retrieve} relevant information from a KB. This contrasts with standard Visual Question Answering (VQA, \cite{antol_vqa_2015}), where questions target the content of the image (e.g. the color of an object or the number of objects) or Knowledge-based VQA (about coarse-grained object categories)~\cite{marino2019ok}, where one can rely on off-the-shelf object detection~\cite{garderes_conceptbert_2020}. 
In KVQAE, both text and image modalities bring useful information that must be combined. Therefore, the task is more broadly related to \textit{Multimodal Information Retrieval} (IR) and \textit{Multimodal Fusion}.

 There are two paradigms for multimodal IR and for multimodal learning more generally: early fusion (data- and feature-level) and late fusion (score- and decision-level)~\cite{kludas_information_2007}. On the one hand, late fusion is more straightforward as both Natural Language Processing and Computer Vision techniques can be applied independently, but it neglects interaction between the modalities. On the other hand, %
the richness of early fusion often comes at the cost of increasing complexity and model parameters. This adds an extra challenge for KVQAE where the two existing datasets are either small (ViQuAE~\cite{sigir2022}) or generated automatically (KVQA~\cite{shah2019kvqa}). To overcome this challenge, we propose to pre-train our fusion model using a multimodal Inverse Cloze Task (ICT). ICT has been introduced in~\cite{lee_latent_2019} to pre-train a neural retriever for textual Question Answering (QA). It consists in considering a sentence as a pseudo-question and its context as a pseudo-relevant passage. It can be seen as a generalization
of the skip-gram objective~\cite{mikolov_distributed_2013}. We extend it by considering images near texts in multimodal documents.

Our main contributions are: (i) Multimodal ICT, a new pre-training method that allows tackling small KVQAE datasets such as ViQuAE; (ii) a multimodal IR framework for KVQAE; (iii) experiments with different neural network architectures, including recently proposed multimodal BERTs.

\section{Related Work}\label{sec:Related Work}

\subsubsection{Dense Retrieval.}
Dense Retrieval is a rapidly evolving field, surveyed in  \cite{lin_pretrained_2021,fan_pre_2021}, with new pre-training tasks, optimizing methods, and variants of the Transformer architecture emerging \cite{ram_learning_2022,hofstatter_efficiently_2021,gao_unsupervised_2021,gao_condenser_2021}.
\cite{lee_latent_2019} were the first to outperform sparse bag-of-words representations such as BM25 with dense representations for QA. Their approach relies on three components: \begin{enumerate*}[label=(\roman*)]
    \item pre-trained language models such as BERT~\cite{devlin2018bert}, which allow to encode the semantic of a sentence in a dense vector;
    \item a contrastive learning objective that optimizes the similarities between questions' and text passages' embeddings (see Section~\ref{sec:Methods}); %
    \item an unsupervised training task, ICT (see Section~\ref{sec:Introduction}).
\end{enumerate*}
\cite{karpukhin2020dense} criticize the latter for being computationally intensive\footnote{\cite{lee_latent_2019} use a batch size of over 4K questions.} and argue that regular sentences are not good surrogates of questions. Instead, they propose DPR, which takes advantage of \begin{enumerate*}[label=(\roman*)]
    \item the heuristic of whether the passage contains the answer to the question to deem it relevant;
    \item unsupervised IR methods such as BM25 to mine hard negatives examples, which proved to be the key of their method’s success.
\end{enumerate*}
We aim at taking advantage of both approaches by \begin{enumerate*}[label=(\roman*)]
    \item pre-training our model on text QA datasets like DPR;
    \item incorporating multimodality into this hopefully-well-initialized model by adapting the ICT of~\cite{lee_latent_2019} to multimodal documents.
\end{enumerate*}

\subsubsection{Multimodal Fusion and Pre-Training.}\label{ssec:Multimodal Fusion}
The success of BERT in NLP~\cite{devlin2018bert}, which relies on the easily-parallelizable Transformer architecture~\cite{vaswani2017attention}, an unsupervised training objective, and a task-agnostic architecture, has concurrently inspired many works in the VQA and cross-modal retrieval fields~\cite{tan_lxmert_2019,lu_vilbert_2019,li_visualbert_2019,su_vl-bert_2020,li_unicoder_2020,chen_uniter_2020}. %
These models are unified under a single framework in~\cite{bugliarello_multimodal_2021} and partly reviewed in~\cite{khan_transformers_2021}. 
All of these models rely on the Transformer architecture, often initialized with a pre-trained BERT, in order to fuse image and text. The training is weakly supervised, based upon image caption datasets such as COCO~\cite{lin_microsoft_2014} or Conceptual Captions~\cite{sharma_conceptual_2018}, and pre-trained object detectors like Faster R-CNN~\cite{ren_faster_2015}. 
\cite{hessel_does_2020} show that these models learn nontrivial interactions between the modalities for VQA.
Multimodal BERTs can be broadly categorized into \textit{single-stream} and \textit{multi-stream}. %
Single-stream models feed both text tokens’ embeddings and image regions’ embeddings to the same Transformer model, relying on the \textit{self-attention} mechanism to fuse them. Instead, in the multi-stream architecture, text and image are first processed by two independent Transformers before using \textit{cross-attention} to fuse the modalities. Both architectures have been shown to perform equally well in~\cite{bugliarello_multimodal_2021}. 
In this work, we use a single-stream model to take advantage of pre-training on text-only (on QA datasets). Also note that, while inspired by these work, we do not use the same training objectives or data, %
which are arguably unsuited for named entities' representations, as explained in the next section.

\subsubsection{Multimodal Information Retrieval and KVQAE.}
Multimodal IR has largely been addressed using late fusion techniques (see~\cite{depeursinge_fusion_2010} for a survey) but we are mostly interested in early fusion techniques in this work. %

\cite{depeursinge_fusion_2010} review first attempts at early fusion. It was then systematically done by concatenating the features of both modalities in a single vector, with a focus on the feature weighting scheme. Concatenation is confronted with the curse of dimensionality as the resulting feature space equals the sum of the
dimensions of each modality’s features. %

\cite{qu_passage_2021} and~\cite{luo_weakly-supervised_2021} concurrently proposed an approach quite similar to ours for Knowledge-based VQA. They adapt DPR~\cite{karpukhin2020dense} by replacing the question encoder with LXMERT~\cite{tan_lxmert_2019}, which allows to fuse the question and image. However, unlike us, they keep the passage encoder based on text-only and use the same pre-training objectives as~\cite{tan_lxmert_2019}, namely Masked Language Modeling, Masked Region Modeling, and Image-Text Matching. We expect that these objectives are suited to learn representations of coarse-grained object categories but not named entities. In other words, they are suited for standard VQA but not KVQAE. For example, Masked Region Modeling relies on an object detector, which is not applicable to KVQAE. While both~\cite{qu_passage_2021} and~\cite{luo_weakly-supervised_2021} experiment on OK-VQA~\cite{marino2019ok}, their results are inconsistent: \cite{qu_passage_2021} show that their model is competitive with a BM25 baseline that takes as input the question and the \textit{human-written} caption of the image while the model of~\cite{luo_weakly-supervised_2021} is outperformed by BM25 with an \textit{automatically-generated} caption. The discrepancies between these works can be explained because they use neither the same KB nor the same evaluation metrics. \cite{guo_unified_2022} also experiment with different multimodal BERTs but dispense passage-level annotation for an end-to-end training of the retriever and answer classifier\footnote{Standard (Knowledge-based) VQA is often treated as a classification task.}.

Although they experiment with KVQA~\cite{shah2019kvqa}, we do not consider the work of~\cite{shevchenko_reasoning_2021,garcia_improving_2021,heo_hypergraph_2022} as their systems take a \textit{human-written} caption as input, which makes the role of the image content unclear.
\cite{shah2019kvqa} follow a late fusion approach at the decision-level. First, they detect and disambiguate the named entity mentions in the question. Then, they rely on a face recognition step as their dataset, KVQA, is restricted to questions about person named entities. Facts from both textually- and visually-detected entities are retrieved from Wikidata\footnote{\url{https://www.wikidata.org/}} and processed by a memory network~\cite{weston2015memory}. In contrast, our work is in line with~\cite{sigir2022}, who use unstructured text from Wikipedia as KB. Unlike~\cite{sigir2022}, who follow a late fusion approach, searching the question and the image independently, %
we aim at a unified representation of the text and image, both on the visual question and KB sides.

\section{Methods}\label{sec:Methods}

In this section, we first formalize our KVQAE framework, then describe the models before diving into the three training stages: (i) DPR for textual Question Answering; (ii) Multimodal Inverse Cloze Task, our main contribution; (iii) Fine-tuning for KVQAE. Finally, we discuss the inference mechanism and implementation details. 

\subsection{Information Retrieval Framework}

In our multimodal setting, both visual questions (from the dataset) and visual passages (from the KB) consist of a text-image pair $(t, i)$, as in Figure~\ref{tab:kvqae_eg}. Our goal is to find the optimal model $E$ to encode adequate representations $\mathbf{q}=E(t_q,i_q)$ and $\mathbf{p}=E(t_p,i_p)$ such that they are close if $(t_p,i_p)$ is relevant for $(t_q,i_q)$ (denoted with the superscripts $(^+)$ and $(^-)$). Search
then boils down to retrieving the $K$ closest visual passages to the visual question. 
When computing the similarity between two vectors, here with the dot product, the objective used throughout all the training stages (§\ref{ssec:Training}) is to minimize the following negative log-likelihood loss for all visual questions in the dataset~\cite{lee_latent_2019,karpukhin2020dense}: $-\log \frac{\exp{(\mathbf{q} \cdot \mathbf{p}^+)}}{\exp{(\mathbf{q} \cdot \mathbf{p}^+)} + \sum_j \exp{(\mathbf{q} \cdot \mathbf{p}_j^-)}}$.
This contrastive objective allows to efficiently utilize passages relevant to other questions in the batch as \textit{in-batch negatives}, since computing the similarity between two vectors is rather inexpensive compared to the forward pass of the whole model. 
We present two different models $E$ in the next section according to their fusion mechanism.

\subsection{Models}\label{ssec:Models}
All of our models take advantage of BERT\footnote{Uncased ‘‘base’’ 12-layers version available at \url{https://huggingface.co}.} %
\cite{devlin2018bert} and CLIP\footnote{With a ResNet-50 backbone~\cite{he_deep_2016}.}~\cite{radford_learning_2021} as building blocks to represent text and image, respectively. BERT is trained for masked language modeling and next sentence prediction on Wikipedia and BooksCorpus~\cite{zhu_aligning_2015}. CLIP has been trained with a contrastive objective in a weakly-supervised manner over 400M image and caption pairs. It has demonstrated better generalization capacities than fully-supervised models and is efficient for KVQAE, as empirically demonstrated in~\cite{sigir2022}. 
We experiment with two different fusion techniques: \model{} and ILF.

Early Cross-Attention fusion (\model{}) is carried out by a single-stream Transformer model like the multimodal BERTs described above (e.g. UNITER~\cite{chen_uniter_2020}). However, instead of relying on a fixed object detector such as Faster R-CNN, we take advantage of CLIP, as motivated above. To enable early fusion, the visual embedding produced by CLIP is projected in the same space as the text using a linear layer with $\mathbf{W}_c \in \mathbb{R}^{c \times d}$ parameters trained from scratch: $\mathbf{e}_c = \mathrm{CLIP}(i) \cdot \mathbf{W}_c$. The resulting embedding is then concatenated with the word embeddings in the sequence dimension, acting as a ‘‘visual token’’. Those embeddings are then fed to the Transformer model, where the attention mechanism should enable interaction between the modalities. The final embedding corresponds to the special \cls{} token: $\mathrm{\model{}}(t,i) = \mathrm{BERT}([t;\mathbf{e}_c])_{\cls{}}$. The Transformer model is first initialized from BERT.

Intermediate Linear Fusion (ILF) introduces an additional  $\mathbf{W}_t \in \mathbb{R}^{d \times d}$ parameters trained from scratch used to simply project the representation of the \cls{} token in the same space as the CLIP embedding before summing the two\footnote{Note that this is equivalent to concatenating both before projecting like $[ \mathrm{BERT}(t)_{\cls{}}; \mathrm{CLIP}(i)] \cdot [\mathbf{W}_t; \mathbf{W}_c]$.}: $\mathrm{ILF}(t,i) = \mathrm{BERT}(t)_{\cls{}} \cdot \mathbf{W}_t  + \mathbf{e}_c$. 

Because both \model{} and ILF produce multimodal representations $\mathbf{q}$ and $\mathbf{p}$, ranking is done directly using their similarity $\mathbf{q} \cdot \mathbf{p}$. As baseline, we follow~\cite{sigir2022} and linearly combine text and image similarities after zero-mean and unit-variance normalization (omitted in the following equation):
\begin{equation}
    \alpha \times \mathrm{BERT}(t_q)_{\cls{}} \cdot  \mathrm{BERT}(t_p)_{\cls{}} + (1-\alpha) \times \cos(\mathrm{CLIP}(i_q),  \mathrm{CLIP}(i_p)) 
\end{equation}
The interpolation hyperparameter $\alpha$ is optimized on the validation set using grid search to maximize Mean Reciprocal Rank. The left term (text similarity) is referred to as DPR in the rest of the paper.

\subsection{Training stages}\label{ssec:Training}

The models are trained sequentially in three stages:

\subsubsection*{Stage 1: DPR for textual Question Answering.}
Leaving visual representations aside, a DPR model is trained starting from the BERT initialization~\cite{karpukhin2020dense}. DPR consists of two BERT encoders: one for the question $t_q$ and one for the text passage $t_p$. %
We use the model pre-trained by~\cite{sigir2022} on TriviaQA, filtered of all questions used in their dataset, ViQuAE. 
They use the KILT~\cite{petroni2020kilt} version of TriviaQA and Wikipedia, which serves as KB in this stage. Each article is then split into disjoint passages of 100 words for text retrieval, while preserving sentence boundaries, and the title of the article is appended to the beginning of each passage. This yields 32M passages, that is $\approx 5.4$ passages per article. Following~\cite{karpukhin2020dense}, 
irrelevant passages (i.e. hard negatives) are mined using BM25~\cite{robertson_okapi_1995}. 

\begin{figure}[t]
    \centering
    \includegraphics[width=.6\textwidth]{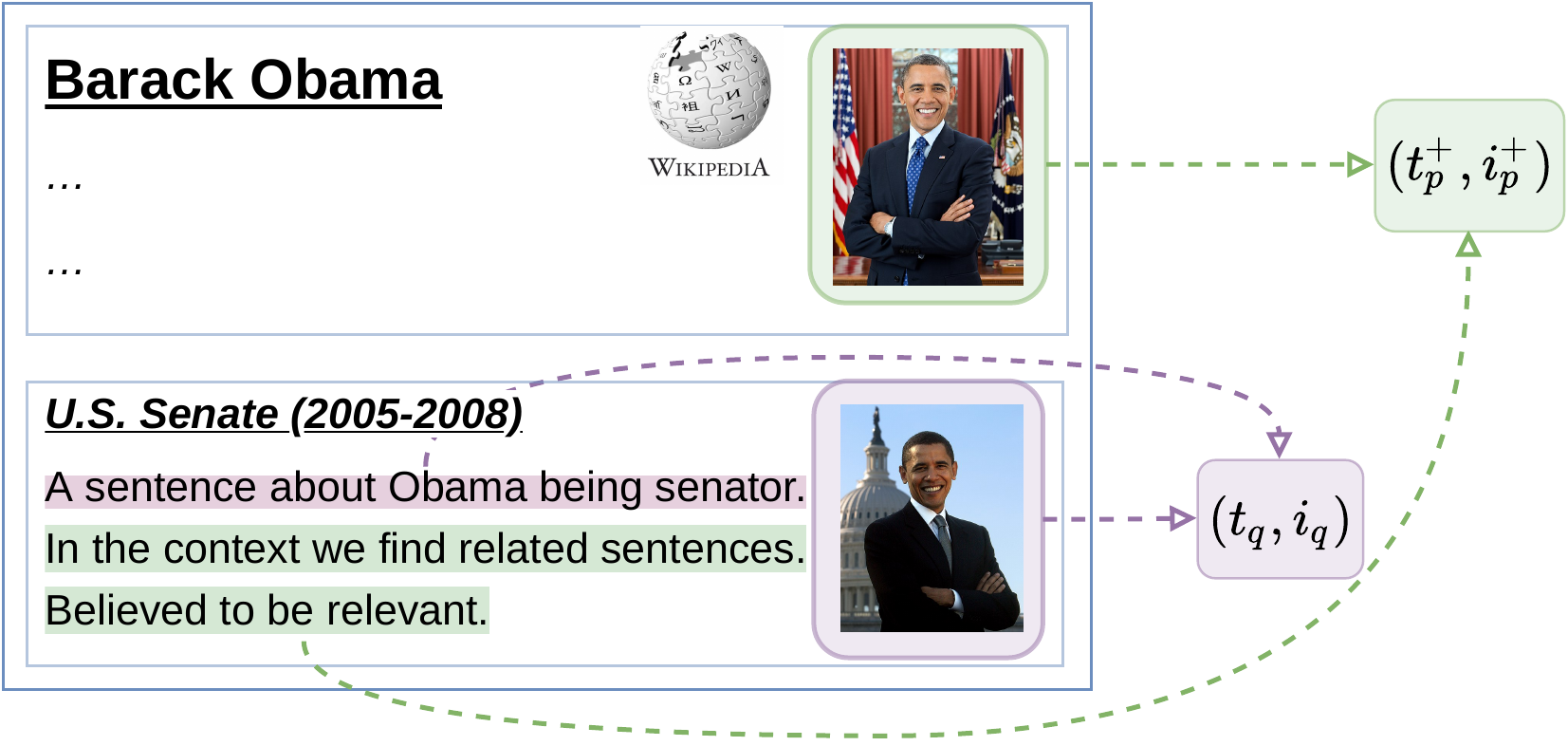}
    \caption{Overview of Multimodal Inverse Cloze Task via Wikipedia/WIT.}%
    \label{fig:ict_from_wiki}
\end{figure}

\subsubsection*{Stage 2: Multimodal Inverse Cloze Task.} 
This is the main contribution of the paper. We propose to extend the ICT of~\cite{lee_latent_2019} to multimodal documents. ICT consists in considering a sentence as a pseudo-question $t_q$ and its context as a pseudo-relevant passage $t_p^+$. Note that the title of the article is appended to the beginning of each passage $t_p$ (as in Stage 1). We extend it using the contextual images of Wikipedia paragraphs for the pseudo-question and the \textit{infobox} image for the passage (see Figure~\ref{fig:ict_from_wiki}). 
\cite{lee_latent_2019} empirically demonstrated that a key success of their approach was to leave the pseudo-question in the relevant passage in 10\% of the training samples so that the model will learn to perform word matching, as lexical overlap is ultimately a very useful feature for retrieval. In our case, however, we argue that it is neither necessary, as the model should be strongly initialized from Stage~1 training on TriviaQA, nor beneficial, as the model could then ignore the image modality. Question and passage encoders pre-trained in Stage~1 are used to initialize the visual question and visual passage encoders, respectively. %

The process is eased thanks to the WIT dataset~\cite{srinivasan_wit_2021}. WIT consists of millions of images with associated text from Wikipedia and Wikimedia Commons in 108 different languages. We are, however, only interested in English for this work. While~\cite{srinivasan_wit_2021} have multiple strategies to find text related to a given Wikipedia image, such as its Commons’ caption, we use only the contextual paragraph as text source in order to mimic the downstream KVQAE setting. The resulting English subset of WIT yields 400K \textit{infobox} images/articles that correspond to 1.2M paragraphs/images. Those 1.2M paragraphs consist of 13.6M sentences, i.e. potential pseudo-questions, which are 26 words long on average. %
Therefore, to stick as close as possible to stages 1 and 3, where passages are up to 100 \textit{words} long, passages consist of \textit{four} sentences. This slightly differs from~\cite{lee_latent_2019} who consider passages of up to 288 \textit{wordpieces}, \textit{prior} to the pseudo-question masking. %

Because both ViQuAE and WIT images are taken from Wikimedia Commons\footnote{\url{https://commons.wikimedia.org/}}, we can estimate from the image URLs that 14\% of ViQuAE images overlap with WIT. This might lead to a bias that we analyze in Section~\ref{ssec:Intrinsic Evaluation}.

Inspired by~\cite{aytar_cross-modal_2017}, to prevent \textit{catastrophic forgetting} and enforce a \textit{modality-invariant} representation of the entities, the last $l$ layers of BERT are frozen during this stage. In this way, we tune only the first, modality-specific layers of ECA, the intuition being to ‘‘replace’’ the text-named entities learned during Stage~1 with the ‘‘visual’’ entities present in the images. ILF fully freezes BERT during this stage, relying only on the $\mathbf{W}_t$ parameters to tune the text representation.
Furthermore, CLIP is systematically frozen throughout all stages. 

We do not have a straightforward way of mining irrelevant visual passages in this stage. In early experiments, we tried to synthesize them by permuting images in the batch: $(t_p^+, i_p^+) \gets (t_p^+,i_p^-)$, but it did not improve the results. %

After filtering corrupted images or images with inappropriate image formats (e.g. .svg) and paragraphs with a single sentence, we end up with 975K paragraphs/images. We refer to it as WIT in the rest of the paper. It is split into train (878K), validation (48K, to tune hyperparameters), and test (48K, as a sanity check) subsets such that there is no overlap between articles.

\subsubsection*{Stage 3: Knowledge-based Visual Question Answering about Named Entities.}

This stage consists in fine-tuning the model on a downstream KVQAE dataset, which provides visual questions $(t_q, i_q)$ and relevant visual passages $(t_p^+, i_p^+)$. Following~\cite{aytar_cross-modal_2017}, all layers of the model are tuned during this stage.

A subtlety of this stage is the selection of irrelevant visual passages $(t_p^-, i_p^-)$. As mentioned in Section~\ref{sec:Related Work}, it was shown to be essential to DPR~\cite{karpukhin2020dense}, and it is more generally important for contrastive learning~\cite{kalantidis_hard_2020}.
In~\cite{sigir2022}, irrelevant passages are mined with BM25 to train DPR. However, we suppose that this is suboptimal for \model{} and ILF as BM25 will only mine textually-plausible passages but not visually-plausible ones. Therefore, we use the system provided by~\cite{sigir2022} to mine irrelevant passages. It is a late-fusion of DPR, ArcFace~\cite{deng_arcface_2019}, CLIP, and ImageNet-ResNet~\cite{he_deep_2016}. This leads to different training setups between DPR (used as a baseline) and our models. However, we have experimented both for DPR and found no significant differences\footnote{Evaluation methods are detailed in Section~\ref{sec:Results}}.

We use the same KB as~\cite{sigir2022}, which is based upon KILT's Wikipedia and Wikidata images of the corresponding entities. It consists of 1.5M articles (thus images/entities) split into 12M passages of at most 100 words as in Stage~1.

Visual questions in ViQuAE are split into train (\numprint{1190}), validation (\numprint{1250}), and test (\numprint{1257}) without overlap between images’ URLs~\cite{sigir2022}.
We do not experiment with KVQA~\cite{shah2019kvqa} for the following reasons: (i) it is generated automatically from Wikidata so our text-based KB has a poor coverage of the answers; (ii) it comprises yes/no questions for which passage relevance cannot be assessed automatically. %

\subsection{Inference}\label{ssec:Inference}

For efficient retrieval, every passage in the KB is embedded along with its corresponding image by the visual passage encoder  beforehand. Given a question grounded in an image, both are embedded by the visual question encoder. Search is then carried out with %
maximum inner product search using Faiss~\cite{johnson_billion-scale_2019}.

\subsection{Implementation Details}\label{ssec:Implementation Details}
Our code is built upon PyTorch~\cite{paszke_pytorch_2019}, Hugging Face’s \texttt{transformers}~\cite{wolf_huggingfaces_2020} and \texttt{datasets}~\cite{lhoest_datasets_2021} (itself wrapping Faiss). It is freely available along with the data and trained models\footnote{\codeurl{}}.

To train \model{}, we use the same hyperparameters as~\cite{sigir2022} for DPR, themselves based upon~\cite{karpukhin2020dense}. In particular, we use a learning rate of  $2 \times 10^{-5}$ along with the Adam optimizer~\cite{kingma_adam_2017}. It is scheduled linearly with 100 and 4 warm-up steps for stages 2 and 3, respectively. However, for ILF, we found, based on the validation set, that it converged faster with a learning rate of  $2 \times 10^{-3}$ and a constant scheduler during Stage~2. 
We believe this is because ILF fully freezes BERT in Stage~2, so it does not require careful scheduling or a small learning rate. 
Dropout~\cite{srivastava_dropout_2014} is applied in BERT and after projecting embedding with $\mathbf{W}_c$ and $\mathbf{W}_t$ with a probability of 0.1 (as in the standard BERT configuration). Likewise, layer normalization~\cite{ba_layer_2016} is applied in BERT and after summing the two embeddings in ILF. Gradients’ norms are clipped at 2.

Models in stages 2 and 3 are trained with a batch size of 512 and 298 visual questions, respectively. %
The success of contrastive learning partly relies on a large number of \textit{in-batch negatives} and, therefore, a large batch size~\cite{qu_rocketqa_2021}.
We found that gradient checkpointing~\cite{chen_training_2016} enables the use of much larger batch sizes for \model{}\footnote{It is not necessary for ILF that fully freezes BERT.}. %
Instead of~\cite{sigir2022} who use \textit{four} NVIDIA V100 GPUs with 32GB of RAM each for a total batch size of 128 questions, we are able to fit a batch of 298 questions (as stated above) in a \textit{single} V100 GPU. Stage~2 takes most of the compute budget, with most models converging after $\approx8$K steps, which takes around three days\footnote{Jean Zay GPUs consume 0.482kW (or 0.259kW after heat recovery) in France, which has an average grid emission factor of 0.0569 kgCO2e/kWh according to \url{https://bilans-ges.ademe.fr/en}.
}. 
Checkpoint selection is made based on the validation \textit{in-batch} Mean Reciprocal Rank, for all stages. In-batch means that only the other visual passages in the batch are ranked and that each visual question is paired with only one relevant visual passage (as during training). 

\section{Results}\label{sec:Results}
The retrieval models are evaluated in two different ways: (i) by computing standard IR metrics on visual passage retrieval; (ii) by feeding retrieved visual passages to a reader module that is tasked with extracting the concise answer to the question, thus achieving KVQAE. Put differently, either evaluate whether the system is able to retrieve a \textit{relevant passage} for the question or whether it is able to \textit{answer} the question. We find both metrics to correlate. Ablation studies are carried out with IR metrics.

ViQuAE is based upon TriviaQA, so it is only distantly supervised: the answer is considered correct if it string-matches the ground truth and, likewise, a passage is deemed relevant if it contains the ground truth\footnote{After standard preprocessing (lowercasing, stripping articles, and punctuation).}. Moreover, Wikipedia aliases of the ground truth are considered to be valid answers.

\subsection{Information Retrieval}\label{ssec:Intrinsic Evaluation}

\begin{table*}[t]

\caption{
IR evaluation on ViQuAE. $l$: Number of frozen layers during Multimodal ICT. Superscripts denote significant differences in Fisher's randomization test with $p\le0.01$. Hits@1 is omitted as it is equivalent to P@1.
}
\centering
\begin{tabular}{llcllll}
\toprule
\textbf{\#}
& \textbf{Model}
& \textbf{Multimodal ICT}
& \textbf{MRR@100}
& \textbf{P@1}
& \textbf{P@20}
& \textbf{Hits@20} \\ 
\midrule
 a &
DPR   &
NA &
32.8\hphantom{$^{bcdef}$} &
22.8\hphantom{$^{bcdef}$} &
16.4\hphantom{$^{bcdef}$} &
61.2\hphantom{$^{bcdef}$} \\
b &
DPR + CLIP &
NA &
34.5$^{a}$\hphantom{$^{cdef}$} &
24.8$^{a}$\hphantom{$^{cdef}$} &
15.8\hphantom{$^{acdef}$} &
61.8\hphantom{$^{acdef}$} \\
c &
 \model{} &
\xmark{}  &
34.6\hphantom{$^{abdef}$} &
25.9$^{a}$\hphantom{$^{bdef}$} &
17.2$^{ab}$\hphantom{$^{def}$} &
61.6\hphantom{$^{abdef}$} \\
d 
& \model{} ($l=6$)&
\cmark{} &
\textbf{37.8}$^{abce}$\hphantom{$^{f}$} &
26.7$^{a}$\hphantom{$^{bcef}$} &
\textbf{19.5}$^{abce}$\hphantom{$^{f}$} &
\textbf{67.6}$^{abce}$\hphantom{$^{f}$} \\
e &
\model{} ($l=0$)&
\cmark{}  &
35.1\hphantom{$^{abcdf}$} &
24.7\hphantom{$^{abcdf}$} &
17.6$^{b}$\hphantom{$^{acdf}$} &
63.7\hphantom{$^{abcdf}$} \\
f &
 ILF ($l=12$)& \cmark{}  &
37.3$^{a}$\hphantom{$^{bcde}$} &
\textbf{26.8}$^{a}$\hphantom{$^{bcde}$} &
19.1$^{abce}$\hphantom{$^{d}$} &
66.9$^{abc}$\hphantom{$^{de}$} \\
\bottomrule
\end{tabular}
\label{tab:ir_viquae_results_yes_triviaqa}
\end{table*}

\begin{figure}[t]
    \centering
\includegraphics[width=\textwidth]{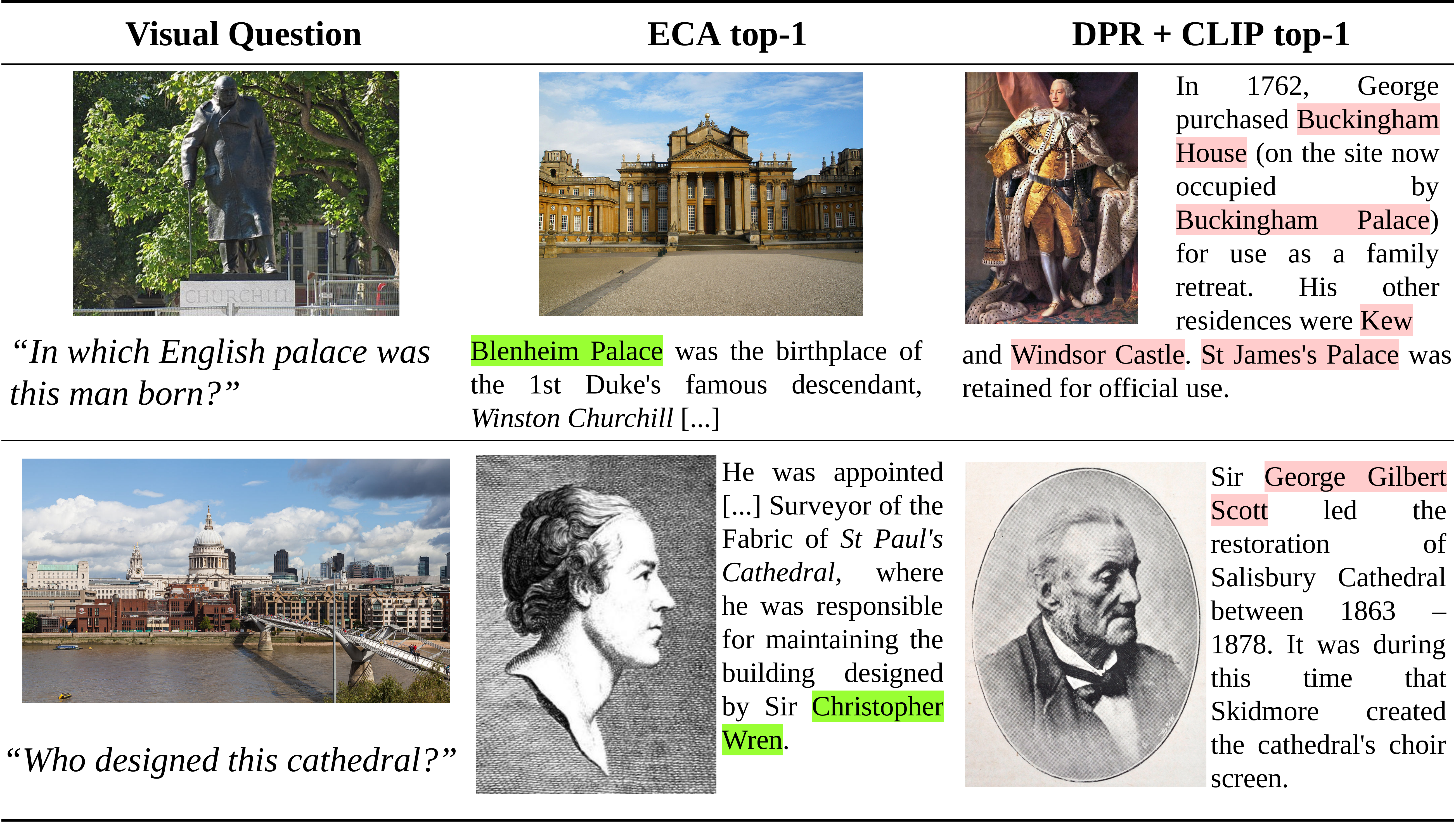}
    \caption{Qualitative examples where \model{} ($l=6$) finds a relevant visual passage in top-1 but late fusion falls behind.} %
    \label{fig:dmr_wins}
\end{figure}

Because of the setting of ViQuAE, it is impossible to get complete coverage of relevant passages. Therefore we do not use any metric based on recall (e.g. R-Precision, mAP, etc.). Instead, we evaluate the models with Precision@K (P@K), Mean Reciprocal Rank (MRR), and Hits@K.  Hits@K is the proportion of questions for which  IR retrieves \textit{at least one} relevant passage in top-K. Statistical significance tests are conducted using Fisher's randomization test~\cite{fisher_design_1937,smucker_comparison_2007}. Metrics and statistical tests are computed with \texttt{ranx}~\cite{bassani_ranx_2022} and are reported in Table~\ref{tab:ir_viquae_results_yes_triviaqa}.

The best models pre-trained with Multimodal ICT (d and f) outperform the text-only (a) and late-fusion (b) baselines on all metrics. Some qualitative examples are shown in Figure~\ref{fig:dmr_wins}. In the first row, we can see evidence of cross-modal interaction between the image depicting Winston Churchill and the passage that mentions him (while being illustrated by a totally different image). In contrast, the late fusion baseline exhibits textual bias by returning a passage that mentions several English palaces (highlighted in red). The same observation can be made for the second row, where St Paul's Cathedral is only mentioned in the relevant passage but not depicted in the contextual image. Cross-modal interactions prove useful in this case because of the heterogeneity of visual depictions: %
Winston Churchill is depicted by a statue in the visual question but by a photograph in the KB.%

We can see that Multimodal ICT is essential to \model{} (c vs. d). Without it, it performs on par with late fusion. We believe this is because of overfitting on the small training set of ViQuAE.
However, we find that fine-tuning on ViQuAE is also essential to \model{}, which exhibits catastrophic forgetting because of the sequential learning setup: indeed, after Stage~2, it falls behind DPR. We see that the freezing technique of~\cite{aytar_cross-modal_2017} helps to prevent catastrophic forgetting to some extent (d vs. e). It is also visible in the upstream WIT pre-training where \model{} achieves  91.6 and 92.9 in-batch MRR on WIT’s test set with $l=6$  and $l=0$, respectively: %
fitting WIT better leads to further forgetting.

Unlike what is suggested by related work (§\ref{ssec:Multimodal Fusion}), we find that the linear fusion model performs on par with the more early, cross-attention based, fusion model (f vs. d). This suggests that the improvement over the late fusion baseline indeed comes from the Multimodal ICT pre-training, which is not very sensitive to the model’s architecture. Moreover, the architecture of ILF allows to fully freeze BERT during Stage~2, which circumvents catastrophic forgetting\footnote{ILF only achieves 87.1 in-batch MRR on WIT’s test set because of the freezing.}. We leave other training strategies (e.g. multi-tasking, using adapters~\cite{houlsby_parameter_2019}) for future work. %

Nothing suggests that \model{} is better on the 14\% of ViQuAE images that overlap with WIT. \model{} is better on the out-of-WIT subset (38.0 vs. 36.5 MRR), but it is the other way around for DPR and late fusion.

\subsection{Reading Comprehension}\label{ssec:Extrinsic Evaluation}

\begin{table}[t]

\caption{
Reading Comprehension evaluation on ViQuAE, averaged over 5 runs of the \textit{reader}. $l$: Number of frozen layers during Multimodal ICT.
}
\centering
\begin{tabular}{ccccc}
\toprule
\textbf{\#} & \textbf{IR Model}
& \textbf{Multimodal ICT}&
   \textbf{Exact Match} &    \textbf{F1} \\ 
\midrule

a & DPR  &
NA &
16.9 $\pm$ 0.4 &    20.1 $\pm$ 0.5 \\

b & DPR + CLIP &
NA & 19.0 $\pm$ 0.4&    22.3 $\pm$ 0.4  \\
c &
 \model{} &
\xmark{}  & 17.7 $\pm$ 0.6 & 21.2 $\pm$ 0.8 \\
d & \model{} ($l=6$)&
\cmark{}  & 20.6 $\pm$ 0.3 &    24.4 $\pm$ 0.2 \\
e &
\model{} ($l=0$)&
\cmark{}  & 20.8 $\pm$  0.8  & 24.3 $\pm$ 0.9\\
f &
 ILF ($l=12$)& \cmark{}  & \textbf{21.3} $\pm$ 0.6 & \textbf{25.4} $\pm$ 0.3 \\
\bottomrule
\end{tabular}
\label{tab:rc_viquae_results_yes_triviaqa}

\end{table}

To extract the answers from the retrieved passages, we keep the same model as~\cite{sigir2022}. It uses the Multi-passage BERT architecture~\cite{wang_multimodal_2019} and is thus based on text only because \textit{once the relevant passage has been retrieved}, the question may be answered without looking at the image. To limit the variations due to training and the number of experiments, we use the model trained by~\cite{sigir2022} off-the-shelf and simply change its input passages. It takes the top-24 passages as input. The model was first trained on TriviaQA (filtered of all questions used in ViQuAE), then fine-tuned on ViQuAE, much like stages 1 and 3. The authors provide \textit{five} different versions of the model that correspond to different random seeds. 

We use Exact Match and F1-score (at the bag-of-words level)  to evaluate the extracted answers.
In Table~\ref{tab:rc_viquae_results_yes_triviaqa} we can verify that more relevant passages indeed lead to better downstream answers. The only difference with the IR evaluation is the role of the freezing technique of~\cite{aytar_cross-modal_2017} (d vs. e), which is less clear here.

\section{Generic vs. Specialized Image Representations}\label{sec:Negative Results}

Numbers reported in the previous section are actually on par with the best results of~\cite{sigir2022}. This is because the latter is based on ArcFace and ImageNet-ResNet, in addition to DPR and CLIP. In particular, \cite{sigir2022} have a heuristic for taking advantage of the face representations provided by ArcFace: they use ArcFace if faces are detected and a combination of CLIP and ImageNet-ResNet otherwise. They show that this method improves retrieval precision for questions about persons (for which face representations are relevant). However, this approach is not scalable for two reasons: (i) there are near 1,000 different entity types in ViQuAE (according to Wikidata’s ontology), and not all can benefit from specialized representations; (ii) combining several representations (e.g. CLIP and ImageNet-ResNet) for the same entity type is computationally expensive and quickly saturates. 
To provide a comparable system to the late fusion of~\cite{sigir2022}, we have tried integrating ArcFace and ImageNet-ResNet in \model{}. However, we have failed to outperform the CLIP-only version of \model{}. Intuitively, we think that \model{} dilutes the specialized representations of ArcFace and is unable to preserve them throughout all twelve layers of BERT. Therefore, in this setting, \model{} is overall on par with late fusion (37.7 vs. 37.9 MRR, not significant) but better on questions about non-persons (39.3 vs. 35.7 MRR), which again suggests that it is unable to exploit ArcFace’s representations.

\section{Conclusion and Perspectives}\label{sec:Discussion}
We have presented a new pre-training method, Multimodal Inverse Cloze Task, for Knowledge-based Visual Question Answering about Named Entities. Multimodal ICT leverages contextual images in multimodal documents to generate pseudo-visual questions. It enables the use of more complex multimodal fusion models than previously proposed late fusion methods. Consequently, our method improves retrieval accuracy over the latter by 10\% relative-MRR, leading to a 9\% relative-F1 improvement in downstream reading comprehension (i.e. answer extraction), on the recently introduced ViQuAE dataset. We believe it is thanks to cross-modal interactions, which are prohibited by late fusion. More precisely, we qualitatively observed that these interactions occurred between the image of the visual question and the text of the KB, which counteracts the heterogeneity of visual depictions. %

We have experimented our pre-training method with two different neural networks architectures: (i) \model{}, which follows recently proposed Multimodal BERTs by fusing modalities Early via Cross-Attention; (ii) ILF, a more standard model that fuses modalities through a linear projection. We found that both perform equally well, unlike in standard VQA and cross-modal retrieval. We argue that it might be because of their difference in training settings, which leads \model{} to catastrophic forgetting. However, further investigations are required. %

While aiming for generic multimodal representations of named entities, we found that integrating specialized representations  in our models, such as ArcFace for faces, was not beneficial. We hypothesize that the studied architectures may be inappropriate but we leave this issue for future studies.

For future work, we think that generalizing Multimodal ICT for re-ranking (processing $(t_q, i_q)$ and $(t_p, i_p)$ simultaneously) and reading comprehension (generating or extracting the answer from $(t_p, i_p)$) is an exciting research lead. Indeed, there is evidence that sharing the same model for IR and reading comprehension, or IR and re-ranking, is beneficial for textual QA~\cite{fun_efficient_2021} and cross-modal retrieval~\cite{geigle_retrieve_2021}, respectively: %
two tasks that closely relate to KVQAE.

\bibliographystyle{splncs04}
\bibliography{Bibliography}

\end{document}